\documentclass[11pt,a4paper]{article}
\usepackage[hyperref]{naaclhlt2018}
\usepackage{times}
\usepackage{latexsym}
\usepackage{graphicx}
\usepackage{stfloats}
\usepackage{url}

\aclfinalcopy 

\title{Frequency vs. Association for Constraint Selection \\
 in Usage-Based Construction Grammar}

\author{Jonathan Dunn \\
 University of Canterbury \\
  Department of Linguistics \\
  {\tt jonathan.dunn@canterbury.ac.nz} }
\date{}

\begin{document}
\maketitle
\begin{abstract}
A usage-based Construction Grammar (CxG) posits that slot-constraints generalize from common exemplar constructions. But what is the best model of constraint generalization? This paper evaluates competing frequency-based and association-based models across eight languages using a metric derived from the Minimum Description Length paradigm. The experiments show that association-based models produce better generalizations across all languages by a significant margin.
\end{abstract}

\section{Learning Slot-Constraints}

The Construction Grammar paradigm (CxG: Langacker, 2008; Goldberg, 2006) represents grammar using a hierarchical inventory of constraint-based \textit{constructions}. In computational terms, a construction is a possibly non-continuous sequence in which each unit satisfies some combination of lexical, syntactic, and semantic constraints (e.g., Chang, et al., 2012; Steels, 2004, 2012, 2017). This paper uses computational modelling to approach the problem of how slot-constraints are learned: do frequency-based or association-based models produce better slot-constraints? How can we evaluate the quality of slot-constraints across an entire grammar in order to make such a comparison possible?

Implementations of CxG such as Fluid Construction Grammar (FCG) and Embodied Construction Grammar (ECG) require the manual specification of constraints using a knowledge representation framework like FrameNet (e.g., Laviola, et al., 2017; Matos, et al., 2017; van Trijp, 2017; Ziem \& Boas, 2017; Dodge, et al., 2017). While these approaches can provide high-quality representations, they cannot model the emergence of slot-constraints because their constraints are \textit{defined} rather than \textit{learned}. We instead follow work that models CxG from a usage-based perspective: first, generating potential constructions given a corpus (Wible \& Tsao, 2010; Forsberg, et al., 2014); second, selecting the optimal set of constructions, where optimality is measured against a test corpus (Dunn, 2017, 2018a). This provides a model of how syntactic constraints are learned.

Recent work has used the Minimum Description Length paradigm (MDL: Rissanen, 1978, 1986; Goldsmith, 2001, 2006) to model the interaction between slot-constraints across an entire grammar as a trade-off between memory and computation. The grammar which selects the best constraints will optimize the balance between memory (the encoding size of all constructions) and computation (the encoding size of a test corpus given the grammar). This operationalizes the idea within usage-based theories of grammar that any representation can be stored in memory but that not all representations are worth storing (c.f., Jackendoff, 2002; O'Donnell, et al., 2011). From a different perspective, some constructions prevent the learning of other constructions (Goldberg, 2011; Goldberg, 2016; Perek \& Goldberg, 2017).

This paper first considers how constructions and slot-constraints can be represented computationally using a data-driven pipeline (Sections 2 \& 3). After describing the data used for the experiments (Section 4), we motivate the contrast between frequency and association (Section 5). The frequency-based and association-based models are described (Sections 6 \& 7), along with a construction extraction algorithm (Section 8). Finally, an MDL approach to grammar quality is motivated (Section 9) and used to evaluate the grammars produced by the two extraction algorithms (Section 10). The experiments show that an association-based model produces better generalizatons for each language, although the degree of difference between the two models varies across languages.

\section{Representing Constructions}

\begin{table*}[t]
\centering
\begin{tabular}{|l|}
\hline
(1a) [SYN:\textsc{noun} --- SEM-SYN:\textsc{transfer[V]} --- SEM-SYN:\textsc{animate[N]} --- SYN:\textsc{noun}] \\
(1b) ``He gave Bill coffee." \\
(1c) ``He gave Bill trouble." \\
(1d) ``Bill sent him letters." \\
(2a) [SYN:\textsc{noun} --- LEX:``give" --- SEM-SYN:\textsc{animate[N]} --- LEX:``a hand"] \\
(2b) ``Bill gave me a hand." \\
\hline
\end{tabular}
\caption{Construction Notation and Examples}
\label{tab:1}
\end{table*}

Following previous work (Dunn, 2017, 2018a), constructions are represented as a sequence of slot-constraints, as in (1a). Slots are separated by dashes and constraints are defined by both type (Syntactic, Joint Semantic-Syntactic, Lexical) and filler (for example: \textsc{noun}, a part-of-speech or \textsc{animate}, a semantic domain). 

The construction in (1a) contains four slots: two with joint semantic-syntactic constraints and two with simple syntactic constraints. The examples in (1b) to (1d) are tokens of the construction in (1a). Lexical constraints, as in (2a), represent idiomatic sentences like (2b). These constructions are context-free because any sequence that satisfies the slot-constraints becomes a token or instance of that construction.

The difficulty of modelling slot-constraints is that constructions can overlap: multiple constructions in the grammar are allowed to represent a single phrase. For example, (2b) is actually a token of both (1a) and (2a). This makes identifying constructions more difficult because reaching the representation in (1a) does not rule out also reaching the representation in (2a). Both could be part of a single speaker's grammar. For this reason we will use the term \textit{extraction} rather than \textit{parsing} to describe the syntactic analysis that is performed by the algorithms described in this paper.

\begin{figure*}[b]
\centering
\includegraphics{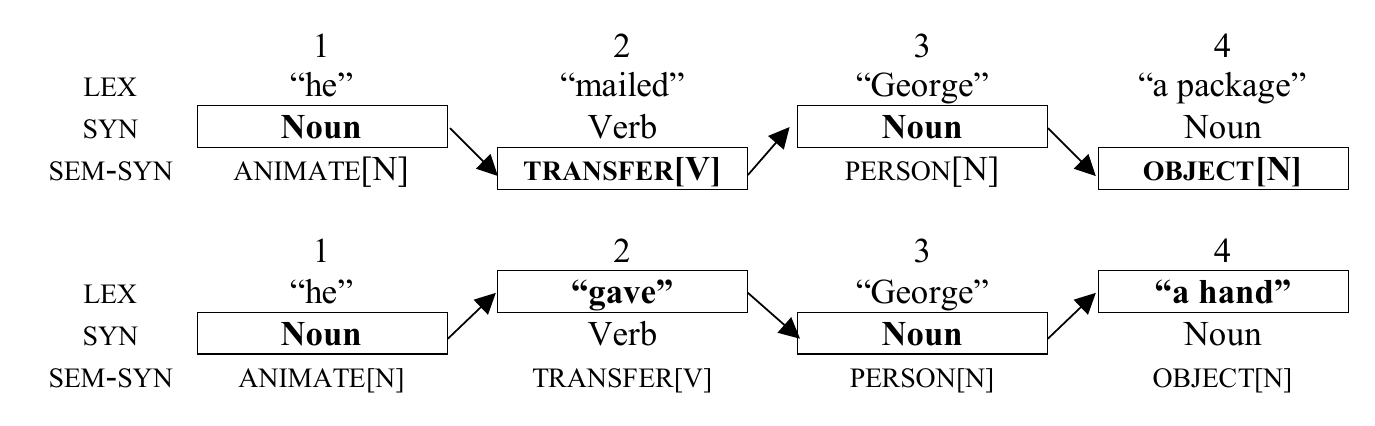}
\caption{General vs. Idiomatic Ditransitive}
\end{figure*}

To illustrate the problem of construction extraction, we can view each slot as a node, with the beginning of a construction the root node (c.f., transition parsing for dependency grammars: Zhang \& Nivre, 2012; Goldberg, et al., 2013). A construction's root can occur anywhere in a sentence. Each slot-constraint is a state, as visualized in Figure 1 with two forms of the ditransitive. There are four possible transitions: \textsc{lex, syn, sem-syn, stop}. In the first example, the slot-constraints are generalized to any transfer verb and any object noun. In the second example, the verb and object slots require idiomatic lexical items. The problem is to find the sequence of slot-constraints that \textit{best} represents the construction. Here, the \textit{best} representation is the most efficient trade-off between memory and computation across an entire grammar.

We first have to develop a pipeline for representing all the possible constraints shown in Figure 1. Such a pipeline provides our hypothesis space: any sequence of constraints that is observed in the training data is a potential construction.

\section{Representing Slot-Constraints}

This section describes the pipeline that is used to represent the hypothesis space of potential constructions. While it is important to take an empirical approach and evaluate aspects of this representation pipeline, the purpose of this paper is not to provide a counter-factual for each component individually (e.g., what type of embeddings or which part-of-speech tags to use). Instead, the two competing approaches are evaluated using the same representation pipeline in order to put such development decisions in the background. Without a pre-defined ontology of concepts and frames, as in knowledge-based CxG, the representation of slot-constraints becomes a difficult problem.

First, lexical constraints use word-forms separated at whitespace; no morphological analysis is included in the pipeline. The lexicon of allowed word-forms is drawn from a background corpus (Section 4), with a frequency threshold to determine inclusion (500 occurrences in corpora of approximately 1 billion words). An example of a lexical slot-constraint is given in (2a), where this particular construction requires the specific words ``give" and ``a hand", as in (2b). 

Second, syntactic representations are drawn from the part-of-speech categories in the Universal POS tagset using the RDRPOS tagger (Petrov, et al., 2012; Nguyen, et al., 2016); this is a pre-defined syntactic ontology. An unsupervised inventory of syntactic units is outside the scope of this paper, although ideally this would also be part of the representation pipeline. An example of a syntactically-defined slot-constraint is given in (2a), in which any noun can fill the subject position. The problem of recursion within slots is discussed further in Section 11.

Third, semantic constraints are defined using a domain dictionary in which each word-form is assigned to a cluster of word-forms. Clusters are based on word embeddings. First, a background corpus for each language is pos-tagged. No word sense disambiguation is used but word-forms are separated by syntactic category (i.e., \textit{table\_verb} is distinct from \textit{table\_noun}). A skip-gram embedding with 500 dimensions is trained for each language (\u{R}eh\r{u}\u{r}ek \& Sojka, 2010). Clusters are then formed by applying x-means to these embeddings (Pelleg \& Moore, 2000). While previous work used k-means to create a fixed number of domains across languages (Dunn, 2018a), x-means generalizes the number of clusters per language. 

These clusters are heterogenous syntactically. Each output cluster is further divided by syntactic category so that each semantic cluster only contains words from a single part-of-speech, allowing joint semantic-syntactic constraints. The number of clusters for each language, shown in Table 2, ranges from 236 (zho) to 487 (por). This variation shows the importance of using x-means for defining semantic constraints instead of k-means with a fixed $k$ across languages.

\begin{table}
\centering
\begin{tabular}{|c|c|c|}
\hline
\textbf{Language} & \textbf{Lexicon Size} & \textbf{N. Clusters} \\
\hline
ara & 57,216 & 315 \\
deu & 43,080 & 305 \\
eng & 47,723 & 385 \\
fra & 46,876 & 326 \\
por & 65,173 & 487 \\
rus & 49,616 & 324 \\
spa & 51,683 & 438 \\
zho & 59,127 & 236 \\
\hline
  \end{tabular}
  \caption{Semantic Clusters by Language}
  \label{tab:1}
\end{table}

\section{Corpora and Data Divisions}

This paper evaluates models on eight languages: Arabic (ara), German (deu), English (eng), French (fra), Portuguese (por), Russian (rus), Spanish (spa), and Chinese (zho). Each language is represented by a large background corpus that is used to (i) train word embeddings, (ii) determine the word-form lexicon, (iii) calculate association measures, and (iv) learn and evaluate CxGs. This section discusses data sources and preparation.

A large portion of the corpus for each language comes from web-crawled data (Baroni, et al., 2009; Majl\c{i}s \& \c{Z}abokrtsk\'{y}, 2012; Benko, 2014; and  data from the CoNLL 2017 Shared Task: Ginter, et al., 2017). Because the goal is to provide a wide representation of each language, this is augmented by legislative texts from the EU and UN (Tiedemann, 2012; Skadi\c{n}\^{s}, et al., 2014), the OpenSubtitles corpus (Tiedemann, 2012), and newspaper texts. The only language-specific pre-processing used is Chinese text segmentation\footnote{Jeiba: \url{https://github.com/fxsjy/jieba}}. 

All punctuation is removed and text converted to lowercase. In order to avoid language-specific assumptions, no sentence splitting is performed. Instead, the corpus is divided into sequences of 100 words that form the main unit of analysis. The corpus is further divided into chunks of 100k samples (for a total of 10 million words per chunk). These chunks are important because the data is randomly divided by chunk as shown in Table 3.

\begin{table}
  \centering
  \begin{tabular}{|ll|}
\hline
\textbf{Function} & \textbf{Num. Words} \\
Word embeddings & Entire dataset \\
Background statistics  & 200 million words \\
Generating potentials & 50 million words \\
Optimizing CxGs & 10 million words \\
Evaluation & 10 million words (x5) \\
\hline
  \end{tabular}
  \caption{Data Divisions}
  \label{tab:1}
\end{table}

We perform CxG learning across four independent folds. Each fold retains the same lexicon and semantic domains, but every other part of the pipeline is repeated: (i) calculating frequency and association statistics for evaluating potential constructions, (ii) generating potential constructions, (iii) searching through the potential constructions using a tabu search (Dunn, 2018a) to optimize the MDL metric (Section 9). 

Each fold produces a single CxG. These CxGs are then merged by concatenation into a single grammar. The idea is that any construction which is productive on a sub-set of the corpus belongs in the final grammar. This final CxG is reduced using horizontal pruning (c.f., Wible \& Tsao, 2010) to remove constructions that are wholly or partially contained within larger constructions. The code for this process is provided as an external resource.

\section{Frequency and Association}

The representation pipeline provides a rich hypothesis space from which to formulate slot-constraints. A usage-based grammar expects that constructions will emerge as common exemplars become entrenched via repeated production and perception. But how do we model \textit{emergence}?

One approach uses frequency: the most common templates (i.e., sequences of constraints) will become a part of the grammar (Bybee, 2006; Arnon \& Snider, 2010; Siyanova-Chanturia, et al., 2011). On the other hand, frequency alone will over-represent very common phrases and we know that less common and even rare constructions remain perfectly grammatical. How do learners acquire rare constructions if they learn using frequency information? A second approach uses association: slot-constraints that occur together more frequently than expected indicate an entrenched construction (Wible \& Tsao, 2010; Forsberg, et al., 2014; c.f., Ellis \& Larsen-Freemen, 2009). An association-based model focuses on frequency relative to specific contexts rather than overall frequency in all contexts.

On the one hand, frequency and association as measures of entrenchment do not need to be mutually exclusive. For example, association measures explicitly depend on frequency information. On the other hand, the purpose of the experiments in this paper is to evaluate competing models of the emergence of slot-constraints against corpus data in order to better understand how CxGs are acquired. It should also be noted that it is not possible to design an association-based algorithm that has no frequency thresholds whatsoever: we need at least a bound on which transitions need to be assigned association values. In the same way, the frequency-based algorithm references some association information; otherwise the number of candidates either will be intractibly large or will include no infrequent forms. Regardless, the algorithms described in Sections 6 and 7 represent implementations of competing hypotheses about the emergence of slot-constraints.

For association, we use the bi-directional $\Delta P$ (Gries, 2013; Dunn, 2018b), with both left-to-right and right-to-left variants. For any two slot-fillers, $X$ and $Y$, $X_P$ indicates that $X$ is present and $X_A$ that $X$ is absent, providing the two direction-specific measures below.
$$\Delta P_{LR} = p(X_P|Y_P) - p(X_P|Y_A)$$
$$\Delta P_{RL} = p(Y_P|X_P) - p(Y_P|X_A)$$
Why not other measures of association? First, the $\Delta P$ was developed for precisely this sort of problem (Ellis, 2007). Second, the $\Delta P$ is bi-directional while other common measures like pointwise mutual information (PMI) average both directions together, thus disguising directional asymmetries. It has been shown that directional association is necessary to describe many linguistic patterns (Gries, 2013). Here we use the maximum directional $\Delta P$ to represent each transition. While a PMI disguises directional differences, this max $\Delta P$ allows each possible transition to be represented by its strongest association value.

\section{Frequency-Based Constraints}

The frequency-based algorithm works in two stages: First, it greedily selects slot-constraints for each sentence by iterating over all adjacent pairs and adding the pair with the highest $\Delta P$ (Table 4). Once all slot-constraints are filled, the second stage extracts constraint n-grams from this sequence ($n$ = 3--6). This approach posits many different boundaries and uses overall frequency across the corpus to prune candidates. $RS$ in Table 4 refers to a sequence of slot-constraints that represents the input sentence; this sequence is complete when every slot in the sentence is represented by a hypothesized constraint. 

\begin{table}
\centering
\begin{tabular}{|l|}
\hline
\textbf{Variables} \\
\hline
$line$ = sequence of units \\
$unit$ = possible slot-constraints: (lex, syn, sem)  \\
$u_i, u_{i+1}$ = two adjacent units \\
$c_i, c_{i+1}$ = constraint types for $u_i, u_{i+1}$ \\
$RS$ = one slot-constraint per unit in line \\
\hline
\textbf{Algorithm} \\
while $RS$ not complete: \\
\hspace{2mm} for $u_i , u_{i+1}$ in line: \\	
\hspace{4mm} for all possible transitions $c_i , c_{i+1}$: \\
\hspace{6mm} if $\Delta P(c_i , c_{i+1})$ is highest available: \\
\hspace{8mm}  add $c_i , c_{i+1}$ to $RS$ \\
\hline
\end{tabular}
\caption{Frequency-Based Selection Algorithm}
\label{tab:1}
\end{table}

This is similar to a template-based view of CxG: each n-gram of slot-constraints is a template. Only the most frequent templates are considered in the MDL stage. On the other hand, it is not tractable to include every sequence of slot-constraints; past work that took such an approach (Dunn, 2017) had to operate on much less data or enforce a series of intermediate frequency thresholds (i.e., per-chunk thresholds). For practical reasons the algorithm in Table 4 references local association between slot-constraints; at its core, however, this is an operationalization of a frequency-centered model of the emergence of slot-constraints.

This frequency-based algorithm uses a fixed frequency threshold. After all candidates are extracted from a corpus, those candidates with an overall frequency below the threshold are pruned. It is difficult to evaluate different thresholds using a grid search approach (as done below with association) because many thresholds produce candidate sets that are too large to evaluate. For purely practical reasons, then, the frequency threshold is fixed. Along these same lines, horizontal pruning removes any candidate that is entirely contained within another candidate, with the larger candidate always remaining and the smaller candidate always pruned. This type of pruning is essential for a frequency-based model because a frequent sequence $A-B-C-D$ will have frequent sub-sequences like $A-B-C$ and $B-C-D$. This nesting is not produced by an association-based model, and so a different pruning strategy is required, as described in Section 7.

\section{Association-Based Constraints}

The association-based algorithm (Table 5) uses the total directional $\Delta P$ (a sum across all transitions) to evaluate potential sequences. To implement this idea, the search follows transitions from one slot-constraint to the next, proceeding left-to-right through the sentence. Any transition below a threshold $\Delta P$ stops that line of the search. This algorithm references local association values when choosing a transition from the current state. It also references global (i.e., construction-wide) association for selecting different paths, rather than using the frequency of specific templates.

Any series of constraints identified by this search whose transitions exceed the $\Delta P$ threshold is added to the candidate stack. At the end of the search, this stack is scored using each candidate's total $\Delta P$ across all transitions. While primarily a transition-based extraction, this approach thus incorporates some global evaluation methods (c.f., Nivre \& McDonald, 2008; Zhang \& Clark, 2008). A grid search for the best $\Delta P$ threshold per language is performed using independent test data.

This association-based algorithm is less influenced by the assumption that co-located slots govern one another's constraints. For example, in reference to Figure 1, the slot filled by a \textsc{noun} in 3 and the slot filled by ``a hand" in 4 have a local transition that is measured using the association between these two representations. Should we instead ignore the relationship between these two objects and focus on the relationship between each object and the verb slot? This algorithm tries to avoid specifying particular templates like this (i.e., a verb-centered frame) by using the global $\Delta P$ evaluation and the thread of associations to draw out these relationships. 

But this raises an interesting empirical question: does the entrenchment of the ditransitive construction predict a higher association between the two object slots whether or not the verb itself is included? Is there a shared effect across all double-object constructions? A beam-search dependency parser could resolve this in a practical sense by simply evaluating more non-local relationships. But does CxG itself predict that such local relationships will be more entrenched because they are present within a single construction? 

\begin{table}
\centering
\begin{tabular}{|l|}
\hline
\textbf{Variables} \\
\hline
$node$ = unit (i.e., word) in line  \\
$startingNode$ = start of potential construction \\
$state$ = type of slot-constraint for node \\
$path$ = route from root to successor states \\
$[c]$ = list of immediate successor states \\
$c_i, c_{i+1}$ = transition to successor constraint \\
$candidateStack$ = plausible constructions \\
$evaluate$ = maximize $\sum \Delta P$ for $c_i, c_{i+1}$ in $path$\\
\hline
\textbf{Main Loop} \\
for each possible startingNode in line: \\
\hspace{2mm} RecursiveSearch(path = startingNode) \\
evaluate candidateStack \\
\hline
\textbf{Recursive Function} \\
RecursiveSearch(path): \\
\hspace{2mm} for $c_i, c_{i+1}$ in $[c]$ from path: \\	
\hspace{4mm} if $\Delta P$ of $c_i, c_{i+1}$ $>$ threshold: \\
\hspace{6mm} add $c_{i+1}$ to path \\
\hspace{6mm} RecursiveSearch(path) \\
\hspace{4mm} else if path is long enough: \\
\hspace{6mm} add to candidateStack \\
\hline
\end{tabular}
\caption{Association-Based Selection Algorithm}
\label{tab:1}
\end{table}
 
\section{Extracting Constructions}

Given a set of candidates (i.e., a possible CxG), we use an additional algorithm to extract those candidates from a corpus in order to evaluate that grammar. The algorithm proceeds left-to-right across each word in the input. For each word, the extractor checks for constructions whose first slot-constraint is satisfied by the current word. Because there are three types of slot-constraints, the extractor must check each constraint type. If the current word satisfies the first slot-constraint, the extractor looks-ahead and tests each successive word until either (i) all slot-constraints are satisfied and a construction match is identified or (ii) a slot-constraint is not satisfied and this portion of the search is terminated. If there is no match, then a particular construction is not present. This algorithm extracts all candidates identified by the above algorithms so that the competing grammars can be evaluated.

\section{Modeling Constraint Quality}

We now have frequency-based and assocation-based models of how slot-constraints emerge from usage. How can we measure the quality of both (i) a set of potential slot-constraints and (ii) an entire CxG? The process of searching over selected slot-constraints using a tabu search (Glover, 1989, 1990) is adopted from previous work (Dunn, 2018a). A tabu search is a meta-level heuristic search that evaluates a number of possible local moves for each turn and then makes the move which produces the best grammar. Importantly, a tabu search allows moves which make the grammar worse in the short-term (with a restricted set of tabu moves) so that the learner can climb out of local optima. Here, each state is a grammar that contains a specific set of constructions. A move changes the current state by adding or removing some constructions. As before, the purpose is not to evaluate counter-factuals for every step in the pipeline because both the frequency-based and association-based models use exactly the same tabu search algorithm.

The MDL metric quantifies the trade-off between memory (operationalized as the encoding size of a grammar) and computation (operationalized as the encoding size of a test corpus given that grammar). A grammar that provides better generalizations will allow the test corpus to be encoded using a smaller number of bits. The metric combines three encoding-based terms: $L_1$ (the cost of encoding the grammar), $L_2\{C\}$ (the cost of encoding pointers to constructions in the grammar), and $L_2\{R\}$ (the cost of encoding linguistic material that is not in the grammar and thus cannot be encoded using a pointer). A pointer here is a partial parse of an utterance that refers to a construction that is already contained in the grammar. 

These terms represent the grammar, the data as described by the grammar, and the data that is not described by the grammar; note that both $L_2$ terms are combined below. In other words, $L_2 (D\mid G)$ is the sum of both $L_2\{C\}$ and $L_2\{R\}$. $D$ in this equation refers to the dataset which is used to evaluate the model. The relationship between these three encoding terms across languages is examined further in Table 7.
$$MDL = \min\limits_{G} \{{L_1(G)+L_2 (D \mid G)}\}$$
Encoding size, in turn, is based on probability: the encoding size of an item, $X$, is measured in bits, below, using the negative log of its probability. We describe how probabilities are estimated later in this section. The basic idea is that more probable constraints should have smaller encoding sizes.
$$L_C (X)=-log_2P(X)$$
According to this model, a construction is only worth remembering if its contribution to decreasing the overall encoding size of the test corpus is smaller than its contribution to the encoding size of the grammar. This is important for CxGs because similar constructions overlap, describing the same sentences in the corpus. Each overlapping construction must be individually represented in the grammar, adding to the $L_1$ term: similar constructions must be encoded separately in $L_1$ but do not improve the encoding of $L_2$. For example, the two constructions in (1a) and (2a) describe the same utterance in (2b). Both of these constructions need to be encoded in the grammar, increasing $L_1$. But encoding only one of them would not increase the regret portion of $L_2$ because the utterance itself can still be encoded using a pointer to the construction that is in the grammar.

The encoding size of a grammar, $L_1$, is the sum of the encoding size of all constructions in that grammar. Each construction is a series of slot-constraints that must be satisfied for a linguistic utterance to be an instance of that construction. For each constraint, two items must be encoded: (i) the constraint type (lexical, semantic, syntactic) and (ii) the filler which defines that constraint. As shown in Table 7, this portion of the MDL metric is quite small given a large dataset.

The cost of (i) is fixed because each representation is considered equally probable: the grammar is not explicitly biased towards syntactic constraints. But the cost of (ii) depends on the type of representation: syntactic units come out of a much smaller inventory, so that any given part-of-speech is more probable and thus easier to encode. For example, if there are 14 parts-of-speech, then the probability of observing one of them is $1/14 =  0.0714$ bits. On the other hand, because there are more lexical items, each word is less probable and thus more expensive to encode.

For example, if there are 50k lexical items, then the probability is $1/50,000 =  0.00002$. In this way, the grammar is allowed to employ item-specific slot-constraints, but doing so increases the encoding cost of the grammar. Here, a syntactic constraint contributes 3.8 bits but a lexical constraint contributes 15.6 bits. Future work will evaluate the impact of probability estimation for slot-fillers, currently done only at the contruction level. The total encoding size of a construction is the accumulated bits required to encode each slot-constraint, where $N_R$ represents the number of representation types (here, 3) and $T_R$ represents the number of possible slot-fillers for that type.
$$\sum_i^{N_{SLOTS}} -log_2(\frac{1}{N_{R_{i}}} ) + -log_2(\frac{1}{T_R})$$
The encoding size of the test corpus, $L_2$, contains two quantities: first, the cost of encoding pointers to constructions in the grammar; second, the cost of encoding on-the-fly any parts of the corpus that cannot be described by the grammar. The cost of encoding pointers is also based on probabilities, so that more probable or common constructions require fewer bits to encode. For example, a construction that occurs 100 times in a corpus of 500k words has a pointer encoding size of 12.28 bits, but a construction that occurs 1,000 times costs only 8.96 bits per use. In this way, the probability of potential constructions influences encoding size. The regret portion of the $L_2$ term is the cost of words which are not covered by constructions in the current grammar. Each of these is encoded on-the-fly (i.e., not remembered): the more unencoded words accumulate, the more each one costs.

There is a close relationship between MDL and Bayesian inference methods (c.f., Barak, et al., 2016; Barak \& Goldberg, 2017; Goldwater, et al., 2009). Information theory describes the relationship between the log probabilities of representations and their encoding size. But it does not estimate the probability of the grammar itself, which here is handled in two ways: First, there is a choice in CxG between different types of representation (\textsc{lex, syn, sem}). This model does not enforce one type, but syntactic constraints are more likely because there are fewer categories. Second, pointers to constructions are assigned probabilities based on their observed frequency; this means that more likely constructions are cheaper to encode and implicitly favored by the model.

The MDL paradigm has previously been applied to phonological structure (Rasin \& Katzir, 2016), to morphological structure (Goldsmith, 2001; 2006), to grammar induction in other contexts (c.f., Solomonoff, 1964; Gr{\"u}nwald, 1996; de Marcken, 1996; c.f., Stolcke, 1994), and even to semantics (c.f., Piantadosi, et al., 2016). This application to CxG incorporates two important properties of usage-based constructions (multiple constraint types and overlapping representations) for which an MDL approach is a good fit.

\section{Does Frequency or Association Produce Better Slot-Constraints?}

\begin{table}
\centering
\begin{tabular}{|cccc|}
\hline
  ~ & Frequency & Association & P \\
\hline
ara & 44.08\% & \textbf{29.45\%} & 0.0001 \\
deu & 52.49\% & \textbf{18.69\%} & 0.0001 \\
eng & 51.80\% & \textbf{23.11\%} & 0.0001 \\
fra & 43.28\% & \textbf{40.52\%} & 0.0037 \\
por & 45.13\% & \textbf{38.91\%} & 0.0137 \\
rus & 54.14\% & \textbf{13.93\%} & 0.0001 \\
spa & 60.34\% & \textbf{26.36\%} & 0.0001 \\
zho & 57.01\% & \textbf{37.96\%} & 0.0030 \\
\hline
\end{tabular}
\caption{Compression Rates by Language with \\ 
Significance of Difference Between Models}
\label{tab:1}
\end{table}

We evaluate the frequency-based and association-based models on the same test sets, with the same hypothesis spaces derived from the same representation pipeline, using the same implementation of the MDL metric. While we have not evaluated counter-factuals for every development decision made within the pipeline, both competing models rely on the same decisions.\footnote{The exact data used is available for download here: \url{https://labbcat.canterbury.ac.nz/download/?jonathandunn/CxG\_Data\_FixedSize}. In addition, the code for the implementation and the grammars themselves are available here: \url{https://github.com/jonathandunn/c2xg/}.}

MDL provides a single metric of a grammar's fit relative to a particular dataset. This metric itself is dependent on each dataset; we thus calculate a baseline encoding score that represents the encoding of the dataset without a grammar and use this to derive a compression metric: $MDL_{CxG}/MDL_{Base}$. The lower this compression metric, the greater the generalizations provided by the CxG. Compression as used in MDL is similar to perplexity within language modelling; the connection is not explored further here except to note that some language models include CxG-like templates (e.g., Gimpel \& Smith, 2011).

The evaluation uses all eight languages in order to provide a cross-linguistic counter-factual: do the generalizations agree across languages? Additionally, we evaluate the models against five independent sets of 10 million words for each language. Table 6 shows the average compression by model for each language across these five test sets. We also report the p-values for a paired t-test (paired by dataset) to ensure that the difference in compression between models is significant. 

Lower compression scores reflect better generalizations; as shown in Table 6, the association-based model out-performs the frequency-based model for every language. In each case the difference between models is significant. The gap and the significance level, however, vary widely across languages. For Russian, there is a gap of 40.21\% compression that is significant below the p = 0.0001 level. But for French and Portuguese that gap is only 2.76\% and 6.22\%, with much larger p-values to match. Association always provides a better model of the emergence of slot-constraints, but for French and Portuguese the two models are much closer together than for other languages.

\begin{table*}[t]
	\centering
	\begin{tabular}{|l|cc|cc|cc|}
		\hline
		
		~ &	$L_1$ ($F$) & $L_1$ ($\Delta P$) & $L_2\{C\}$ ($F$) & $L_2\{C\}$ ($\Delta P$) & $L_2\{R\}$ ($F$) & $L_2\{R\}$ ($\Delta P$) \\
		\hline
		ara & 0.43\% & 1.25\% & 82.14\% & 68.65\% & 17.43\% & 30.10\% \\
		deu & 0.50\% & 1.56\% & 89.32\% & 93.42\% & 10.17\% & 05.01\% \\
		eng & 0.57\% & 1.44\% & 93.22\% & 98.04\% & 06.21\% & 00.53\% \\
		fra & 0.44\% & 0.77\% & 93.08\% & 64.09\% & 06.48\% & 35.14\% \\
		por & 0.39\% & 0.27\% & 96.72\%	& 25.00\% & 02.89\% & 74.73\% \\
		rus & 0.42\% & 1.35\% & 66.37\% & 94.87\% & 33.21\% & 03.78\% \\
		spa & 0.36\% & 0.81\% & 99.59\% & 82.24\% & 00.06\% & 16.95\% \\
		zho & 0.25\% & 0.37\% & 92.24\% & 96.92\% & 07.51\% & 02.71\% \\

		\hline
	\end{tabular}
	\caption{Break-down of MDL metric by relative proportion of the overall score}
	\label{tab:1}
\end{table*}

The frequency-based model represents what Goldberg calls \textit{conservatism via entrenchment}, the idea that learners are more willing to over-generalize infrequent forms (Goldberg, 2016). In other words, the problem with a frequency-based model is that it does not allow for creative (and thus infrequent) uses of common forms. The more frequent a particular form is, the less likely that form will allow competing representations to emerge. But language is infinitely creative and this model blocks too many emerging constructions.

The association-based model, on the other hand, allows for the emergence of less familiar constructions: a series of transitions between slot-constraints is permitted if it is relatively highly associated, and infrequent forms are more open to forming new associations. This is the advantage of a directional measure like the $\Delta P$. Assume there are two constraints, $A$ and $B$, in which $A$ is very common but $B$ is rare. The PMI, by averaging, would disguise any association from $B$ to $A$. But the $\Delta P$ allows such new associations to emerge from a limited number of observations. Frequency alone pre-empts less common representations.

We take a closer look at cross-linguistic patterns in Table 7 by breaking down the MDL metric into its three terms: $L_1$, or the contribution of the complexity of the grammar; $L_2\{C\}$, or the contribution of encoded constructions to the final encoding cost; and $L_2\{R\}$, or the contribution of missing constructions to the final encoding cost. Each term is represented by its percentage of the MDL metric for that dataset. Thus, while the frequency-based models have a higher MDL score overall, Table 7 focuses on how that score is distributed across terms. These percentages are averaged across all five test sets for each language for each model.

First, we see that the grammars themselves ($L_1$) account for a small percentage of the overall metric. The great majority of the MDL score comes from the encoding of pointers or references of constructions in the grammar in order to represent the dataset. A smaller percentage comes from encoding errors (i.e., parts of the dataset that cannot be represented using known constructions).

Second, however, there are important variations across languages and model-types. For French and Portuguese, the two languages with the least difference between frequency-based and association-based models, the association-based models have significantly higher regret encodings ($L_2\{R\}$). In other words, the overall fit of the association-based models for these languages is not nearly as good: only 64.09\% (fra) and 25.00\% (por) of the association-based model's MDL score comes from correctly encoded constructions. This suggests that the association-based model provides relatively poor grammars for French and Portuguese, rather than that the frequency-based model provides relatively good grammars. At the same time, this relative distribution of the MDL metric disguises the fact that the overall compression of the association-based model remains better for these two languages than the frequency-based model.

Could we evaluate usage-based grammars without relying on MDL? An alternate approach to evaluating the balance of memory and computation when learning syntactic structures involves psycholinguistically-annotated datasets (c.f., Luong, et al., 2015) or qualitative distinctions such as optional/obligatory arguments (c.f., Bergen, et al., 2013). These approaches are not as comprehensive as the work described in this paper because they cover, in effect, a small sub-set of possible constructions. Yet their interpretation in respect to individual cognition is more straight-forward.

\section{Remaining Problems}

This section offers a brief discussion of an important remaining challenge: how does a grammar treat non-contiguous constructions? A first option is that a CxG assumes a CFG that provides a notion of constituency: a noun phrase, for example, could be taken as a single slot-filler regardless of its internal structure. From a usage-based perspective, this is potentially problematic: Is there a separate syntactic mechanism for constituents that is based on different capabilities than the mechanism for learning slot-constraints?  

A second option is that constituents are a form of purely-syntactic construction that can fill slots within larger constructions. This simpler type of construction would be learned using the same mechanisms as other constructions, but restricted to only syntactic constraints. Two difficulties arise: First, a constituent has a \textit{head} which categorizes it. Thus, if a constituent were categorized as a \textsc{noun}, it could fill any slot in a larger construction that was categorized to accept a \textsc{noun}. But how do we categorize a construction? Does one single slot in a construction act as the head? The second difficulty is that constructions can overlap within a sentence, as with (1a) and (2a) above. But a constituency grammar cannot allow such ill-defined segmentations.

A third option, taken here implicitly, is to allow partially-filled slots or unfilled slots: rather than posit a constituent with a categorized head as a slot-filler, we could allow a specific head along with unspecified material to fill a slot. For example, assume that the ditransitive in (1a) has ``my uncle's wife" in the recipient slot. We could use a constituency grammar to treat this whole phrase as a single \textit{NP}; but we could also allow  ``wife" to satisfy the slot-constraint on its own and treat its modifiers as under-specified material.

\nocite{*}
\bibliography{NAACL_19}
\bibliographystyle{acl_natbib}

\end{document}